\documentclass{article}
\usepackage{spconf,amsmath,graphicx,hyperref}
\usepackage{amssymb}
\usepackage{booktabs}
\usepackage{multirow}
\usepackage{capt-of}
\usepackage{cite}
\usepackage{amsmath,amssymb,amsfonts}
\usepackage{algorithmic}
\usepackage{graphicx}
\usepackage{textcomp}
\usepackage{xcolor}
\usepackage[table]{xcolor}
\usepackage{xcolor}
\usepackage[normalem]{ulem}
\usepackage{soul}
\useunder{\uline}{\ul}{}
\usepackage{float}
\usepackage{bm}  

\title{GRC-Net: Global Representation Consistency Network \\for Unsupervised Multimodal Anomaly Detection}
\name{Seyoung Jeong$^{1}$, Jong Pil Yun$^{2,3}$, Sang Jun Lee$^{1*}$}
\address{
$^{1}$Jeonbuk National University,
Jeonju, Republic of Korea\\
$^{2}$Korea Institute of Industrial Technology
(KITECH), Incheon, Republic of Korea\\
$^{3}$Chung-Ang University, Seoul, Republic of Korea\\}
\begin{document}
%
\maketitle
\begin{abstract}
Automated quality inspection is essential for ensuring product reliability in manufacturing.
While image-based methods effectively capture appearance-related defects, these methods are limited in detecting structural and geometric anomalies, motivating multimodal approaches incorporating 3D information.
However, existing methods mainly rely on local patch-level representations, which often lead to unstable reconstruction errors even in normal regions.
To address this limitation, we propose GRC-Net, which integrates a global-attention MLP to enforce global representation consistency across patch embeddings with a stable reconstruction module to improve reconstruction stability.
The proposed method captures holistic contextual information through a global token and suppresses reconstruction noise by minimizing discrepancies between original and predicted embeddings. 
Experiments on MVTec 3D-AD and Eyecandies demonstrate that GRC-Net consistently outperforms existing methods at both image and pixel levels. 
Qualitative results further demonstrate reduced reconstruction errors in normal regions and more distinct reconstruction differences between normal and anomalous regions.
\end{abstract}

\vspace{-1mm}
\vspace{-2mm}
\section{Introduction}
\label{sec:intro}
In manufacturing, surface quality inspection is essential for product reliability. However, manual inspection is often inconsistent and inefficient, while diverse defect types and high labeling costs make large-scale defect data collection difficult. 
Therefore, unsupervised anomaly detection using only normal samples has been widely studied, with RGB-based methods serving as a practical and cost-effective solution for industrial inspection.~\cite{tien2023revisiting, yu2021fastflow, lei2023pyramidflow}
However, RGB-based methods rely on surface color and texture information, limiting their ability to detect geometric anomalies such as subtle object deformations~\cite{horwitz2023back}. 
In addition, lighting variations, shadows, and reflections often alter pixel values and cause false detections in normal products.
Accordingly, multimodal anomaly detection methods incorporating 3D information have gained increasing attention for detecting geometric anomalies and improving robustness to environmental variations~\cite{bergmann2023anomaly}.

Recent studies have proposed various multimodal anomaly detection approaches~\cite{wang2025m3dm, zavrtanik2024cheating, zavrtanik2024keep}. 
M3DM~\cite{wang2025m3dm} stores modality-specific features in memory banks and compares them with test features for anomaly detection. However, this leads to high memory consumption in multimodal settings, while misaligned feature spaces often destabilize distance-based comparisons. 
CFM~\cite{costanzino2024multimodal} learns cross-modal feature reconstruction from normal data, but has difficulty reconstructing inherent variations such as surface curvature and material diversity in real-world manufacturing environments.

We propose the global representation consistency network (GRC-Net), a reconstruction-based multimodal anomaly detection method using RGB images and 3D point clouds. 
GRC-Net incorporates a global attention MLP (GA-MLP) to generate a global token that captures the intrinsic variability of normal samples through global representation consistency across patch embeddings. 
A stable reconstruction module (SRM) further improves reconstruction stability and enlarges the reconstruction gap between normal and anomalous regions.
We evaluate GRC-Net on the MVTec 3D-AD~\cite{bergmann2021mvtec} and Eyecandies~\cite{bonfiglioli2022eyecandies} benchmarks. 
Experimental results demonstrate that the proposed method outperforms existing multimodal anomaly detection methods, particularly in image-level AUROC. 
The main contributions of this study are as follows:
\begin{itemize}  
    \item We propose a GA-MLP that generates a global token from feature embeddings of normal data and enforces global representation consistency to predict features of the counterpart modality.
    \item We propose a SRM that suppresses noise in anomaly maps of normal data and enhances the reconstruction gap between normal and anomalous regions.
    \item We conduct extensive experiments on MVTec 3D-AD and Eyecandies, two representative benchmarks for unsupervised multimodal anomaly detection. 
    Our method achieves strong performance on both datasets.
\end{itemize}

\begin{figure*}[tb!]
  \centering
  \includegraphics[width=1.0\linewidth]{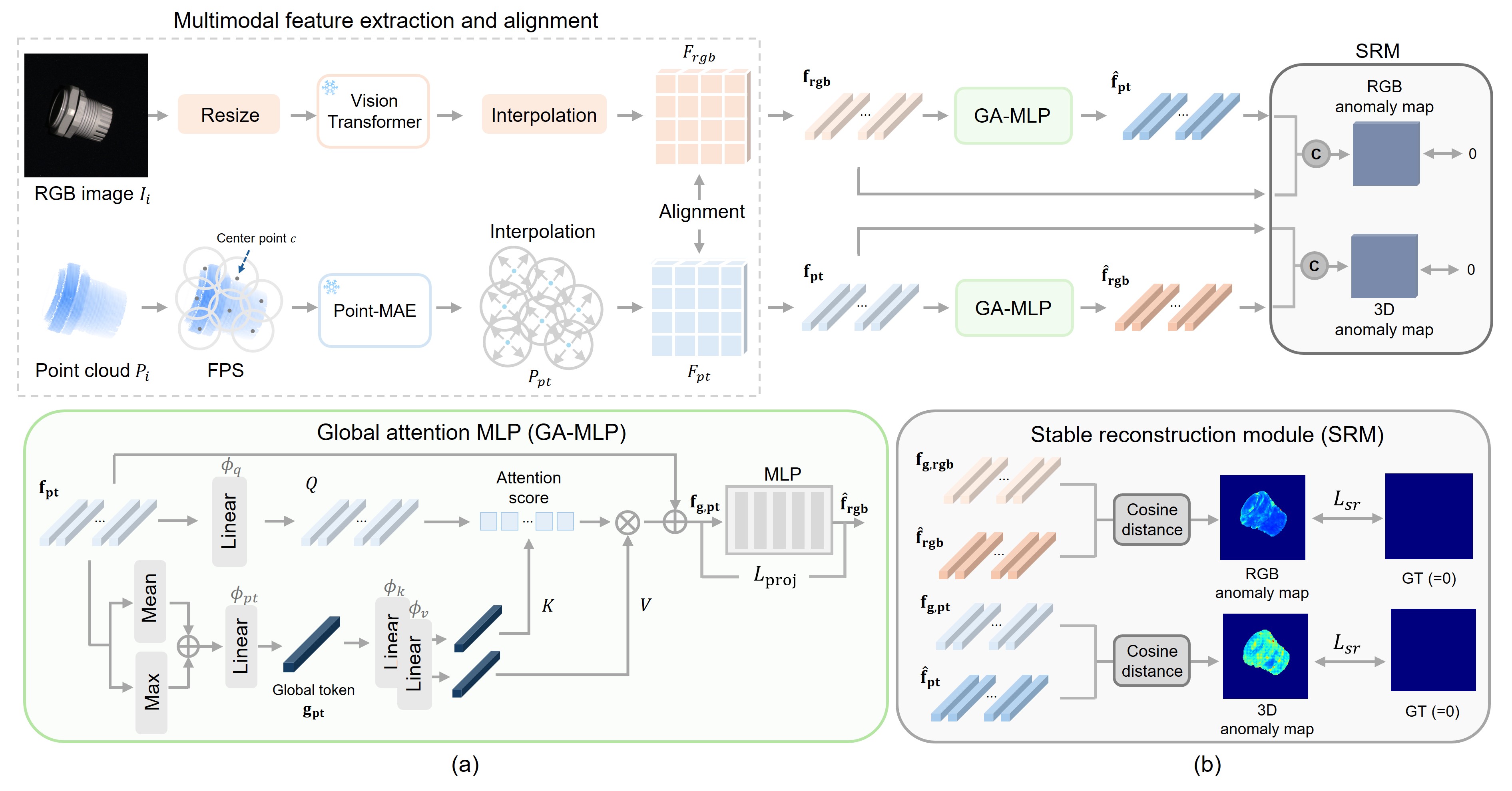}
  \vspace{-8.5mm}
  \caption{Overview of the proposed GRC-Net architecture. 
  The framework incorporates a GA-MLP to capture global contextual relationships and a SRM to suppress unstable reconstruction errors.}
  \label{fig:1}
\end{figure*}

\section{METHODOLOGY}
\label{sec:method}
We propose GRC-Net, a multimodal anomaly detection framework designed to improve reconstruction stability and clarify the reconstruction gap between normal and anomalous region.
The overall architecture is shown in Figure~\ref{fig:1}.
Feature embeddings are first extracted using pretrained modality-specific encoders, and global tokens are generated to enforce global representation consistency across patch embeddings. 
Reconstruction is then performed using globally informed embeddings, while the SRM suppresses reconstruction noise and promotes accurate reconstruction of normal data. 
During inference, modality-specific anomaly maps are obtained from reconstruction errors and combined through pixel-wise multiplication to produce the final anomaly map. 

\subsection{Multimodal feature extraction and alignment}
We extract and align features from RGB images and point clouds using pretrained encoders. Given an RGB image $I_i$ of size $224 \times 224 \times 3$, a DINO-pretrained ViT-B/8~\cite{dosovitskiy2020image, caron2021emerging} on ImageNet~\cite{deng2009imagenet} extracts a $28 \times 28 \times 768$ feature map, which is upsampled by bilinear interpolation to obtain $F_{\mathrm{rgb}}$ of size $224 \times 224 \times 768$.
For a point cloud $P_i$ of size $N \times 3$, we use Point-MAE~\cite{pang2023masked} pretrained on ShapeNet~\cite{chang2015shapenet}. 
The point cloud is divided into 1,024 groups via farthest point sampling, with 32 points per group, and Point-MAE extracts a 1,152-dimensional feature for each representative center point $c$. 
Distance-based interpolation is then applied to obtain $P_{\mathrm{pt}}$, which consists of $N$ interpolated 1,152-dimensional feature vectors. 
Using the known pixel correspondence of each 3D point, $P_{\mathrm{pt}}$ is projected onto the image plane to form $F_{\mathrm{pt}}$ of size $224 \times 224 \times 1152$.
Finally, the pixel-wise aligned feature maps $F_{\mathrm{rgb}}$ and $F_{\mathrm{pt}}$ are obtained, and the corresponding pixel-level feature vectors $\bm{f}{\mathrm{rgb}}$ and $\bm{f}{\mathrm{pt}}$ are used as network inputs.

\subsection{Global attention MLP}
In multimodal reconstruction-based anomaly detection, capturing global contextual information is essential for modeling the intrinsic variability of normal samples. 
To enhance reconstruction stability by promoting global representation consistency, we propose a GA-MLP that injects a global token summarizing the feature distribution of each modality.
The architecture of GA-MLP for the processing of $\bm{f}_{\mathrm{pt}}$ is presented in Fig.~\ref{fig:1}. (a), and the same GA-MLP is also applied to the RGB feature representation $\bm{f}_{\mathrm{rgb}}$.
Specifically, we generate a global token $\bm{g}_{\mathrm{rgb}}$ and $\bm{g}_{\mathrm{pt}}$ using the modality-specific feature vectors $\bm{f}_{\mathrm{rgb}}$ and $\bm{f}_{\mathrm{pt}}$.
The equation is defined as follows:
\begin{equation}
\label{eq:1}
\begin{aligned}
\bm{g}_{\mathrm{rgb}} &=
\phi_{\mathrm{rgb}}\Big(
\mathrm{concat}\big(
\mathrm{Mean}(\bm{f}_{\mathrm{rgb}}),
\mathrm{Max}(\bm{f}_{\mathrm{rgb}})
\big)
\Big), \\[4pt]
\bm{g}_{\mathrm{pt}} &=
\phi_{\mathrm{pt}}\Big(
\mathrm{concat}\big(
\mathrm{Mean}(\bm{f}_{\mathrm{pt}}),
\mathrm{Max}(\bm{f}_{\mathrm{pt}})
\big)
\Big).
\end{aligned}
\end{equation}
Mean pooling and max pooling are applied over all patches of each sample to summarize global context and local saliency, respectively.
The resulting vectors are concatenated along the channel dimension and projected back to the original feature dimension to form the final global token.

Subsequently, all patch tokens are used as queries, while the global token is projected into key–value pairs, allowing each patch to align with the global representation and enforce global representation consistency before being fed into the MLP.
Specifically, the global embeddings $\bm{f}_{g,\mathrm{rgb}}$ and $\bm{f}_{g,\mathrm{pt}}$ are defined as follows:

\begin{equation}
\small
\label{eq:2}
\bm{f}_{g,\mathrm{rgb}}
=
\bm{f}_{\mathrm{rgb}}
+
\mathrm{softmax}
\left(
\frac{
\phi^{\mathrm{rgb}}_{q}(\bm{f}_{\mathrm{rgb}})
\,
\phi^{\mathrm{rgb}}_{k}(\bm{g}_{\mathrm{rgb}})^{\top}
}{\sqrt{C_{\mathrm{rgb}}}}
\right)
\phi^{\mathrm{rgb}}_{v}(\bm{g}_{\mathrm{rgb}}),
\end{equation}

\begin{equation}
\label{eq:3}
\small
\bm{f}_{g,\mathrm{pt}}
=
\bm{f}_{\mathrm{pt}}
+
\mathrm{softmax}
\left(
\frac{
\phi^{\mathrm{pt}}_{q}(\bm{f}_{\mathrm{pt}})
\,
\phi^{\mathrm{pt}}_{k}(\bm{g}_{\mathrm{pt}})^{\top}
}{\sqrt{C_{\mathrm{pt}}}}
\right)
\phi^{\mathrm{pt}}_{v}(\bm{g}_{\mathrm{pt}}),
\end{equation}
In Eq.~(\ref{eq:3}),  $\phi_q(\cdot)$, $\phi_k(\cdot)$, and $\phi_v(\cdot)$ are implemented as modality-specific learnable linear layers with identical input and output dimensions, mapping features to the query, key, and value spaces, respectively, and $C_{\mathrm{rgb}}$ and $C_{\mathrm{pt}}$ denote the channel dimensions of the RGB and point-cloud feature embeddings.
The proposed mechanism reduces visual bias and local noise from patch-level embeddings while improving representation consistency by sharing global patterns. 
The enhanced patch embeddings are then projected into the other modality through a deep nonlinear MLP, enabling more stable and semantically aligned cross-modal mapping while capturing the natural variability of normal samples.

\subsection{Stable reconstruciton module}
Existing reconstruction-based anomaly detection methods assume complete reconstruction of normal data, but this assumption is often violated, resulting high anomaly scores for normal patches. 
We therefore align patch embeddings across RGB images and point clouds through bidirectional feature mapping while reducing reconstruction errors in normal data.

To effectively learn the bidirectional mapping structure, we jointly employ a cross-modal projection loss for semantic alignment between modalities and a stable reconstruction loss to stabilize reconstruction errors on normal data.
The structure of the SRM is presented in Fig.~\ref{fig:1}. (b).
First, we define a cosine-based projection loss that maximizes the directional similarity between predicted features and their corresponding ground-truth features by projecting the patch embeddings $\bm{f}_{g,\mathrm{rgb}}$ and $\bm{f}_{g,\mathrm{pt}}$ of each modality into the feature space of the other modality.
The equation is defined as follows:
\begin{equation}
\small
\label{eq:4}
L^{i}_{\mathrm{proj}}
=
\left(
1 -
\frac{
f^{i}_{g,\mathrm{rgb}} \cdot \hat{f}^{i}_{\mathrm{pt}}
}{
\|f^{i}_{g,\mathrm{rgb}}\|\,
\|\hat{f}^{i}_{\mathrm{pt}}\|
}
\right)
+
\left(
1 -
\frac{
f^{i}_{g,\mathrm{pt}} \cdot \hat{f}^{i}_{\mathrm{rgb}}
}{
\|f^{i}_{g,\mathrm{pt}}\|\,
\|\hat{f}^{i}_{\mathrm{rgb}}\|
}
\right).
\end{equation}
In Eq.~(\ref{eq:4}), $\hat{f}^{i}_{\mathrm{rgb}}$ and $\hat{f}^{i}_{\mathrm{pt}}$ denote the predicted features at pixel $i$, obtained by projecting the patch embeddings of the opposite modality through the corresponding GA-MLP, where the loss is computed independently for each pixel.

However, to compensate for magnitude differences that cannot be captured by cosine-based alignment, we define a stable reconstruction Loss that directly minimizes the normalized difference between the predicted embedding and the ground-truth embedding.
The equation is defined as follows:
\begin{equation}
\small
\label{eq:recon_error}
d^{i}_{m}
=
\left\|
\frac{f^{i}_{g,m}}{\|f^{i}_{g,m}\|}
-
\frac{\hat{f}^{i}_{m}}{\|\hat{f}^{i}_{m}\|}
\right\|_{2},
\quad m \in \{\mathrm{rgb}, \mathrm{pt}\}.
\end{equation}

\begin{equation}
\small
\label{eq:stable_recon_loss}
\begin{aligned}
L_{sr}
&=
\frac{1}{N}
\sum_{i=1}^{N}
\Big(
(d^{i}_{\mathrm{rgb}})^{2}
+
(d^{i}_{\mathrm{pt}})^{2}
\Big).
\end{aligned}
\end{equation}

\begin{equation}
\small
\label{eq:total_loss}
L
=
L_{\mathrm{proj}}^{\mathrm{rgb}}
+
L_{\mathrm{proj}}^{\mathrm{pt}}
+
\alpha \, L_{\mathrm{sr}}.
\end{equation}
In Eq.~(\ref{eq:total_loss}), $\alpha$ denotes the weight of the loss term and is set to 0.2 in our experiments.
By jointly optimizing the projection and reconstruction objectives, the proposed loss formulation stabilizes reconstruction quality for normal data and reduces false anomaly responses caused by unstable embeddings.

\vspace{-1.5mm}

\section{Experiments}
\label{sec:exp}
\noindent\textbf{Experimental details.}
We evaluate the proposed method on MVTec 3D-AD~\cite{bergmann2021mvtec} and Eyecandies~\cite{bonfiglioli2022eyecandies}, standard benchmarks for unsupervised multimodal anomaly detection that provide pixel-wise aligned RGB images and point clouds.
Performance is evaluated using AUROC and AUPRO. 
I-AUROC measures image-level anomaly detection, while P-AUROC evaluates pixel-level anomaly localization. 
AUPRO measures the overlap between predicted and ground-truth anomalous regions, with the threshold set to 0.3. All experiments were conducted on an NVIDIA H100 GPU, and the model was trained for 250 epochs using Adam with a learning rate of 0.001.


\begin{table}[b]
\centering
\caption{
Comparison of quantitative results on the MVTec 3D-AD and Eyecandies datasets.
The highest scores are shown in bold, and the second-highest scores are underlined.
``--'' denotes results not reported in the original paper.
}
\label{tab:comparison}
\resizebox{\linewidth}{!}{
\begin{tabular}{l|ccc|ccc}
\hline
\multirow{2}{*}{Method}
& \multicolumn{3}{c|}{MVTec 3D-AD}
& \multicolumn{3}{c}{Eyecandies} \\
\cline{2-7}
& I-AUROC & P-AUROC & AUPRO
& I-AUROC & P-AUROC & AUPRO \\ 
\hline

AST~\cite{rudolph2023asymmetric}
& 0.937 & 0.976 & 0.944
& 0.780 & 0.902 & 0.744 \\

M3DM~\cite{wang2025m3dm}
& 0.945 & 0.992 & 0.964
& \underline{0.897} & \underline{0.977} & 0.882 \\

EasyNet~\cite{chen2023easynet}
& 0.926 & - & 0.821
& 0.869 & - & - \\

CFM~\cite{costanzino2024multimodal}
& 0.954 & \underline{0.993} & 0.971
& 0.881 & 0.974 & \underline{0.887} \\

2M3DF~\cite{asad20252m3df}
& \underline{0.966} & - & \underline{0.975}
& - & - & - \\

Ours
& \textbf{0.972} & \textbf{0.998} & \textbf{0.976}
& \textbf{0.918} & \textbf{0.982} & \textbf{0.888} \\

\hline
\end{tabular}
}
\label{tab:1}
\end{table}

\begin{figure}[t]
  \centering
  \includegraphics[width=\linewidth]{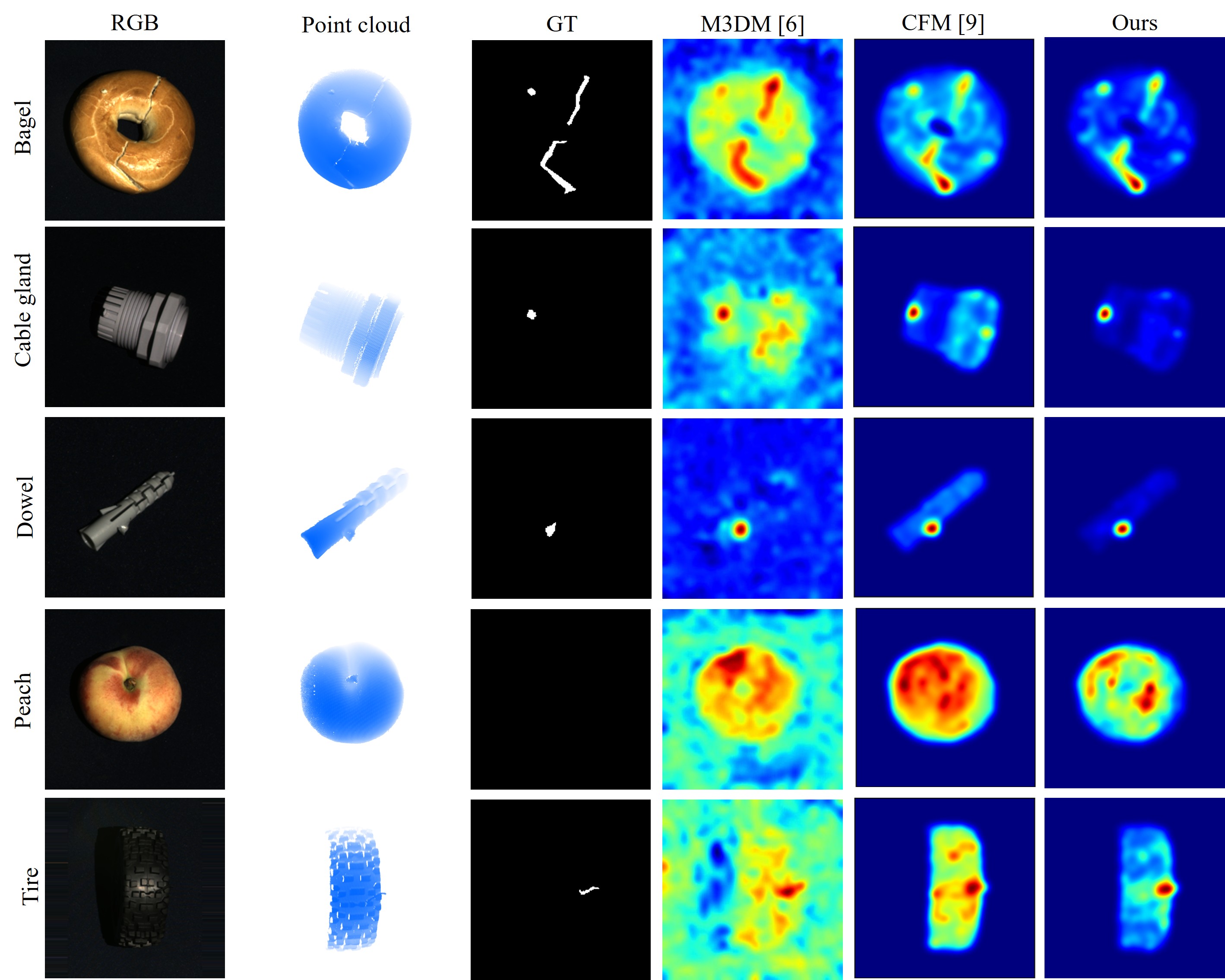}
  \vspace{-7mm}
  \caption{Qualitative comparison of anomaly localization results on the MVTec 3D-AD dataset.
}
  \label{fig:2}
\end{figure}

\subsection{Experimental results}
We present quantitative and qualitative comparisons with previous methods on the MVTec 3D-AD and Eyecandies datasets. 
Table~\ref{tab:1} compares performance in terms of I-AUROC, P-AUROC, and AUPRO. 
Overall, GRC-Net achieves strong performance across both datasets, with particularly high results in image-level and pixel-level AUROC. 
These results demonstrate that the GA-MLP and SRM effectively suppress spurious anomaly responses in normal regions and improve distinction between normal and anomalous areas.

\begin{figure}[t]
  \centering
  \includegraphics[width=\linewidth]{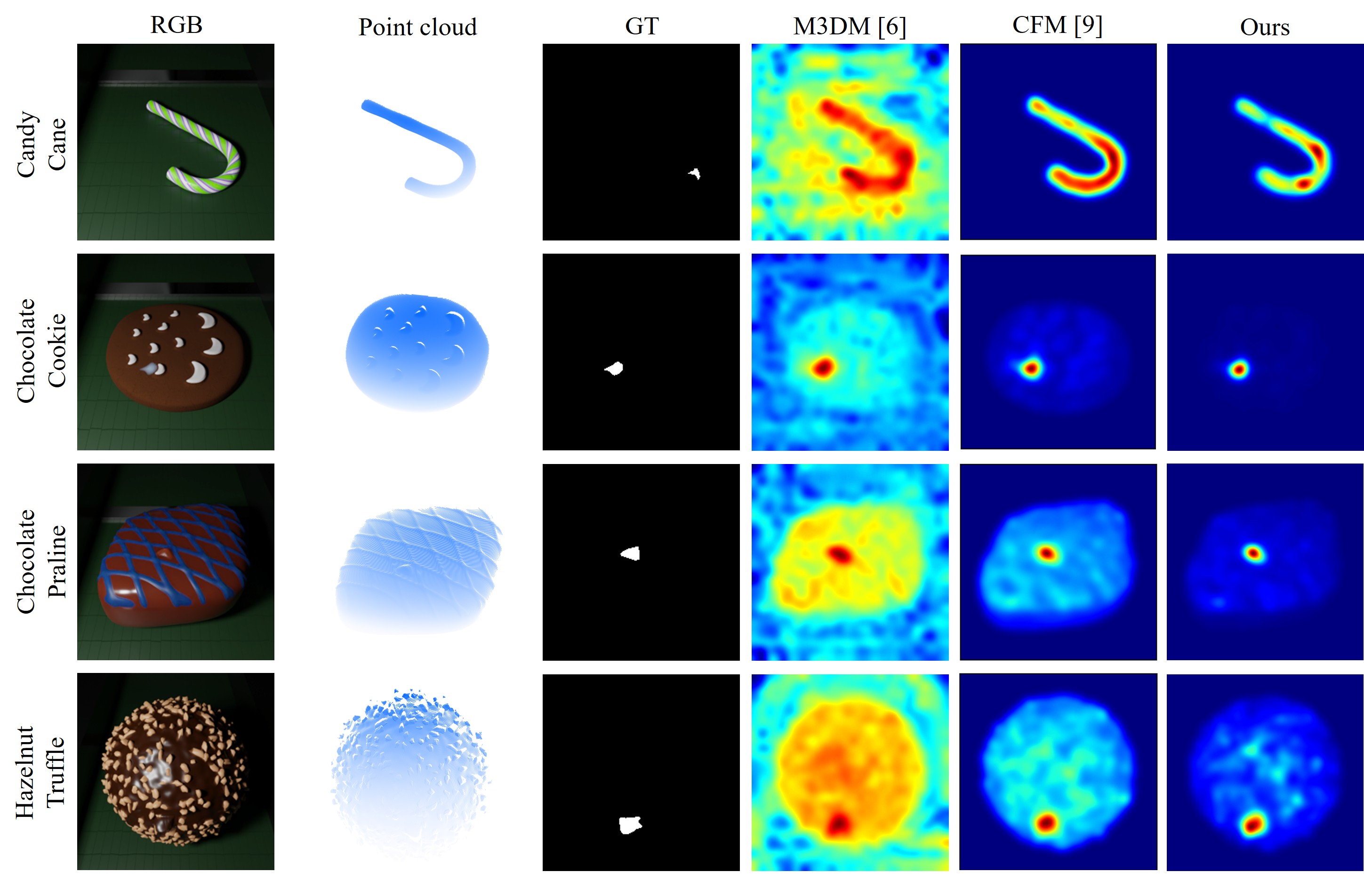}
  \vspace{-8mm}
  \caption{Qualitative comparison of anomaly localization results on the Eyecandies dataset.
}
  \label{fig:3}
\end{figure}

Qualitative comparisons of anomaly maps on MVTec 3D-AD are presented in Fig.~\ref{fig:2}. 
Existing methods often generate noisy responses on irregular or curved surfaces, such as cable gland, peach, and tire. 
In contrast, the proposed method suppresses high anomaly scores in normal regions and more clearly localizes true anomalies. 
In particular, the stable reconstruction loss reduces noisy responses caused by illumination and surface variations, resulting in more refined anomaly maps. 
Fig.~\ref{fig:3} presents qualitative comparisons of anomaly maps on the Eyecandies dataset.
M3DM assigns high anomaly scores to large normal regions, while CFM reduces some noise but still shows false responses and blurred boundaries. In contrast, the proposed method suppresses noise in normal regions and generates anomaly maps that better align with the ground truth. These results demonstrate that the GA-MLP and SRM improve global representation consistency and reconstruction stability.

\subsection{Ablation study}
We conduct ablation studies to evaluate the contributions of GA-MLP and SRM. 
Table~\ref{tab:3} reports I-AUROC, P-AUROC, and AUPRO on MVTec 3D-AD and Eyecandies. 
Applying GA-MLP or SRM individually improves performance over the baseline, while their combination achieves the best results on both datasets. 
These results confirm that GA-MLP enhances global representation consistency and SRM improves reconstruction stability and anomaly localization.
We also analyze the effect of the SRM loss coefficient $\alpha$. 
As shown in Table~\ref{tab:4}, the best performance on both datasets is obtained at $\alpha=0.2$. 
Smaller values provide insufficient regularization, while larger values overly emphasize reconstruction stability and reduce anomaly sensitivity. 
Therefore, $\alpha$ is fixed at 0.2 for all experiments.

\begin{table}[t]
\small
\centering
\setlength{\tabcolsep}{3pt}
\renewcommand{\arraystretch}{1.2}

\caption{Ablation study on the effectiveness of GA-MLP and SRM.}
\vspace{2mm}
\resizebox{\columnwidth}{!}{%
\begin{tabular}{
>{\centering\arraybackslash}m{1.2cm}|
>{\centering\arraybackslash}m{1.2cm}|ccc|ccc}
\hline
\multirow{2}{*}{GA-MLP} 
& \multirow{2}{*}{SRM} 
& \multicolumn{3}{c|}{MVTec 3D-AD} 
& \multicolumn{3}{c}{Eyecandies} \\ 
\cline{3-8}
& 
& I-AUROC & P-AUROC & AUPRO 
& I-AUROC & P-AUROC & AUPRO \\ 
\hline
& & 0.954 & 0.993 & 0.971 & 0.881 & 0.974 & 0.887 \\
\checkmark & & 0.965 & 0.995 & 0.974 & 0.909 & 0.980 & \textbf{0.888} \\
& \checkmark & 0.967 & 0.995 & \textbf{0.976} & 0.917 & 0.981 & \textbf{0.888} \\
\checkmark & \checkmark & \textbf{0.972} & \textbf{0.998} & \textbf{0.976}
& \textbf{0.918} & \textbf{0.982} & \textbf{0.888} \\
\hline
\end{tabular}%
}
\label{tab:3}

\vspace{3mm}

\caption{Ablation study on the effect of the SR loss weight $\alpha$ in the SRM.}
\vspace{2mm}
\resizebox{\columnwidth}{!}{%
\begin{tabular}{
>{\centering\arraybackslash}m{1.0cm}|ccc|ccc}
\hline
\multirow[c]{2}{*}{$\alpha$} 
& \multicolumn{3}{c|}{MVTec 3D-AD} 
& \multicolumn{3}{c}{Eyecandies} \\ 
\cline{2-7}
& I-AUROC & P-AUROC & AUPRO
& I-AUROC & P-AUROC & AUPRO \\ 
\hline
0.1 & 0.968 & 0.995 & 0.975 & 0.915 & 0.981 & 0.883 \\
0.2 & \textbf{0.972} & \textbf{0.998} & \textbf{0.976}
& \textbf{0.918} & \textbf{0.982} & \textbf{0.888} \\
0.3 & 0.964 & 0.995 & 0.974 & 0.917 & 0.980 & 0.881 \\
0.4 & 0.966 & 0.995 & 0.975 & 0.916 & 0.980 & 0.883 \\
\hline
\end{tabular}%
}
\label{tab:4}

\end{table}

\section*{Conclusion}
\label{sec:conclusion}
In this paper, we propose GRC-Net, a multimodal anomaly detection framework using RGB images and point clouds. 
GRC-Net integrates GA-MLP to enforce global representation consistency and SRM to improve reconstruction stability, reducing reconstruction noise in normal regions and enhancing the reconstruction gap between normal and anomalous regions. 
Experiments on MVTec 3D-AD and Eyecandies demonstrate improved image-level and pixel-level anomaly detection performance over existing methods, while qualitative results confirm more accurate anomaly localization.




\bibliographystyle{IEEEbib}
\bibliography{strings,refs}

\end{document}